\pdfoutput=1
\documentclass[letterpaper, 10 pt, conference]{ieeeconf}  

\IEEEoverridecommandlockouts                              

\usepackage{hyperref}       
\usepackage{url}            
\usepackage{booktabs}       
\usepackage{amsfonts}       
\usepackage{nicefrac}       
\usepackage{microtype}      
\usepackage[dvipsnames]{xcolor}
\usepackage{graphicx}
\usepackage{bm}
\usepackage{amsmath}
\usepackage{multirow}
\usepackage{threeparttable}
\usepackage{adjustbox}
\usepackage{eqparbox}
\usepackage{arydshln}
\usepackage{wrapfig}

\usepackage{graphicx}
\usepackage{amsmath,amssymb} 
\usepackage{tabularx}
\usepackage{multirow}
\usepackage{algorithm}
\usepackage{algpseudocode}
\usepackage{float}
\usepackage{pgfplots}
\usepackage{caption}
\usepackage{subcaption}

\title{\LARGE \bf
Domain Adaptation in 3D Object Detection with Gradual Batch Alternation Training
}

\author{Mrigank Rochan$^{1}$, Xingxin Chen$^{2}$, Alaap Grandhi$^{2}$, Eduardo R. Corral-Soto$^{2}$, and Bingbing Liu$^{2}$ 
 	\thanks{$^{1}$Mrigank Rochan is with the Department of Computer Science, University of Saskatchewan  
		{\tt\small mrochan@cs.usask.ca}. Part of the work done while at Huawei Noah's Ark Lab.}%
	\thanks{$^{2}$Authors are with Huawei Noah's Ark Lab. Corresponding author is Bingbing Liu {\tt\small liu.bingbing@huawei.com}.}%
}

\begin{document}

\maketitle
\thispagestyle{empty}
\pagestyle{empty}

\begin{abstract}
We consider the problem of domain adaptation in LiDAR-based 3D object detection. Towards this, we propose a simple yet effective training strategy called Gradual Batch Alternation that can adapt from a large labeled source domain to an insufficiently labeled target domain. The idea is to initiate the training with the batch of samples from the source and target domain data in an alternate fashion, but then gradually reduce the amount of the source domain data over time as the training progresses. This way the model slowly shifts towards the target domain and eventually better adapt to it. The domain adaptation experiments for 3D object detection on four benchmark autonomous driving datasets, namely ONCE, PandaSet, Waymo, and nuScenes, demonstrate significant performance gains over prior arts and strong baselines.

\end{abstract}

\section{Introduction}
Accurate 3D object detection is essential to develop robust LiDAR-based perception systems in autonomous driving. The goal in 3D object detection is to localize and classify the objects in 3D point clouds captured by LiDAR sensors mounted on autonomous vehicles. Since 3D object detection has garnered increasing attention of researchers, we are witnessing promising progress and advancements in its solutions \cite{yin2021center,lang2019pointpillars,shi2020points,shi2020pv,yan2018second}. This success is primarily fueled by advances in deep learning and publicly available large-scale human-labeled LiDAR datasets (e.g., nuScenes \cite{caesar2020nuscenes} and Waymo \cite{sun2020scalability}) for 3D object detection \cite{yang2021st3d}. 

\begin{figure}[t] 
	\centering
	\includegraphics[width=0.42\textwidth]{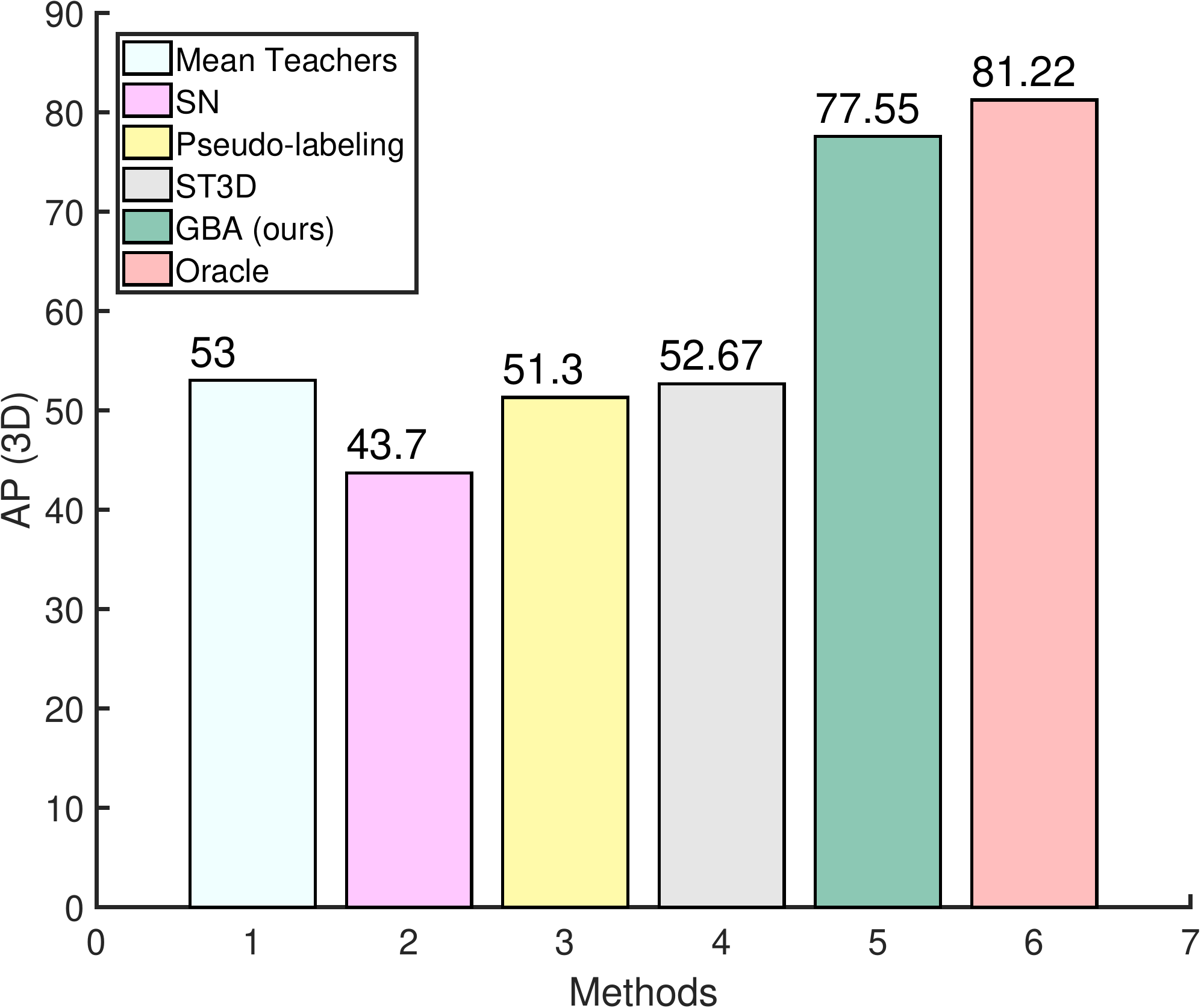} 
    \caption{Comparison of different 3D object detection domain adaptation methods on Waymo $\rightarrow$ nuScenes. Existing domain adaptation methods (e.g., Mean Teachers \cite{tarvainen2017mean}, SN \cite{wang2020train}, Pseudo-labeling \cite{caine2021pseudo}, and ST3D \cite{yang2021st3d}) are essentially unsupervised. Although these methods are promising,  they perform very inferior to Oracle (i.e., fully supervised) which inhibits their adoption in practice. In light of this, we propose \texttt{GBA}, a domain adaptation technique that uses a limited number of labeled target domain samples to significantly reduce the performance gap with Oracle.}
    \label{fig:intro}
\end{figure}

But, a major limitation of the existing methods in 3D object detection is their inability to perform well on a different dataset (target domain) than the dataset (source domain) the method was trained on. This is mainly due to domain shifts likely from variations in LiDAR sensors, environment conditions, and geographic locations \cite{yang2021st3d,caine2021pseudo}. The degradation in the  performance of perception systems in another domain is greatly inhibiting the expansion of this cutting-edge technology. Our experiments also show that when a state-of-the-art supervised 3D object detection model (CenterPoint \cite{yin2021center}) trained on ONCE \cite{mao2021once} or PandaSet \cite{xiao2021pandaset} or Waymo \cite{sun2020scalability} is directly evaluated on nuScenes \cite{caesar2020nuscenes}, the model performance drops drastically (see Source Only results in Table \ref{table:main-results}). As a consequence, there is a pressing need to develop techniques that help LiDAR perception algorithms such as 3D object detection generalize well in novel target domains and make them more robust and reliable for wider deployment. Therefore, there is a stream of recent research in 3D object detection that focuses on domain adaptation methods (e.g., SN \cite{wang2020train}, Pseudo-labeling \cite{caine2021pseudo}, and ST3D \cite{yang2021st3d}). Despite improvement in generalization and results in a new target domain, these methods are still very inferior (by over $28$\% in AP\textsubscript{3D}) compared to the fully-supervised (Oracle) approach which we also illustrate in Fig. \ref{fig:intro} (see the first four bars vs. the last bar).  

In this paper, we aim to minimize the performance gap with the Oracle in the target domain and explore domain adaptation with partial supervision for 3D object detection. In partial supervision, we assume that we have access to a limited number of labeled samples (and no unlabeled samples) from the target domain together with a fully labeled source domain. Since autonomous driving is a safety-critical technology, we believe it is reasonable and practical for autonomous vehicle companies to invest a small amount of capital in labeling a limited number of samples (i.e., point clouds) from the target domain as it can possibly help achieve higher accuracy and strongly compete with the Oracle. Note that our setting is slightly different from semi-supervised domain adaptation where a large set of unlabeled target domain samples are also provided in addition to a small set of labeled target samples and a fully labeled source domain data \cite{saito2019semi}.

We propose Gradual Batch Alternation ($\texttt{GBA}$), a training procedure for domain adaptation with partial supervision and demonstrate its usefulness in adapting a 3D object detector to a new target domain. The main idea in $\texttt{GBA}$ is to smoothly transition the model to the target domain. In $\texttt{GBA}$, we begin to train the model with the full source and the limited labeled target domain data by alternately feeding a batch of source and target samples in each training step, but then gradually reduce the amount of source domain data as the training advances. In the early stages of the training, the amount of source domain data is much more than the target domain data, however, in the later stages the amount of source domain data gradually decrease and become less dominant than the target domain data. In every training stage, the training steps consist of alternate processing of batch of samples from the source and target data as long as possible, but when we run out of samples for alternation from a domain, we continue to train the model with the remaining samples of the larger domain in that stage. Since we train the model on a sequence of data distributions of the source and target domain data where at each point (i.e., training stage) a smaller amount of source data (from the previous stage) is used together with the fixed amount of target domain data, the model learns from data distributions that are increasingly helpful \cite{bengio2009curriculum,xu2021gradual}. Because the training data distribution in $\texttt{GBA}$ gets closer to the target domain over time, the model is able to better adapt to the target domain.

We conduct extensive experiments on benchmark autonomous driving datasets that indicate the strong domain adaptation performance of our approach $\texttt{GBA}$ for 3D object detection. $\texttt{GBA}$ outperforms the prior methods by a large margin and is also able to reduce the performance gap (to less than $4$\%) with the fully supervised Oracle model in the target domain (see the two bars on the right in Fig. \ref{fig:intro}).

To summarize, our contributions are as follows:
\begin{itemize}
	\item We highlight the importance of domain adaptation in 3D object detection and discuss the limitations of the existing methods. 
	\item We propose domain adaptation with partial supervision in the target domain, which is largely unexplored.
	\item We propose the Gradual Batch Alternation ($\texttt{GBA}$) training scheme for domain adaptation in LiDAR-based 3D object detection.
	\item We perform extensive experiments on benchmark autonomous driving datasets for domain adaptation in 3D object detection which demonstrate the superiority of our approach over the existing methods and strong baselines.
\end{itemize}

\section{Related Work}
\subsection{LiDAR 3D Object Detection}
Given a 3D point cloud as input, the 3D object detection task consists of detecting, classifying, and regressing the 3D box parameters of scene objects such as vehicles and pedestrians. LiDAR-based 3D object detection methods can be roughly classified into three family of classes:  1) Point-based methods, 2) Voxel-based methods, and more recently,  3) Hybrid methods that use both, voxelization and point-based techniques.

Point-based methods receive an unordered and unstructured list of 3D points as an input to detect objects and regress 3D bounding boxes. PointNet~\cite{qi2017pointnet}, PointNet++ \cite{qi2017pointnet++}, and its variants introduce successful strategies to process unordered, variable-length lists of 3D input raw point clouds that mainly consist of MLPs and transformations. Methods such as F-PointNet~\cite{qi2018frustum} and PointRCNN \cite{shi2019pointrcnn} adopt PointNet to learn point-level features for generation of region of interest and 3D bounding box regression and their further refinement.  

Projection-based methods (e.g., \cite{chen2017multi,corral2020understanding}) project the 3D point clouds onto an image plane and then detect objects in 3D using 2D convolutions.

Voxel-based methods cluster the input 3D point cloud into regular 3D bins referred to as voxels that are then processed by 3D convolution layers (e.g., VoxelNet \cite{zhou2017voxelnet}) or into pillars for 2D convolution (e.g., PointPillars \cite{lang2019pointpillars}).

The CenterPoint method \cite{yin2021center} utilizes either VoxelNet or Pointpillars as backbone to generate a bird's eye view feature map representation. It then uses 2D CNN detection heads to estimate object centers and perform 3D box parameter regression using features centered at these points. CenterPoints produced state-of-the art vehicle and pedestrian 3D detection results on both the Waymo and nuScenes datasets. For this reason, we select CenterPoint as the 3D object detection model in our method.

More recent methods such as the PV-RCNN~\cite{shi2020pv} combine voxelization and point-based mechanisms to improve the accuracy of 3D object detection. The recent PDA~\cite{hu2022point} method addresses point sampling deficiencies by taking into account the point density within the voxels into the feature learning.

\subsection{Domain Adaptation in LiDAR Perception}
A key contributor to the success of state-of-the-art LiDAR 3D object detection methods (such as CenterPoint) is the availability of large-scale labeled training data from a given domain. However, these methods do not perform equally well on samples (i.e., point clouds) from another domain that may have different LiDAR sensors or representing a new geographic location. This lack of generalization is due to the domain shifts between the domains. An obvious solution is to manually label a huge amount of LiDAR point clouds from the new domain for supervised learning. However, this process is slow and costly.

As a result, some recent research in 3D object detection focuses on domain adaptation. They propose either unsupervised (e.g., ST3D \cite{yang2021st3d}, Pseudo-labeling \cite{caine2021pseudo}, and SPG \cite{xu2021spg}) or weakly supervised (e.g., SN \cite{wang2020train}) domain adaptation methods. There are some methods that transform LiDAR point clouds into images and then perform domain adaptation \cite{corral-soto2021lcp,rochan2022udass}. Although these methods show improvement in generalization, the performance gap with the fully supervised Oracle remains substantial. In this work, we aim to reduce this performance gap and build trustworthy 3D object detection model by improving their ability to generalize well.

Another possible solution to reduce the performance drop due to domain shifts is to collect a small number of labeled samples from the new domain and then perform fine-tuning~\cite{bengio2012deep, goodfellow2016deep,corral2020understanding}. We argue that this is a viable solution for autonomous driving companies since a small investment can help develop safer and reliable solutions. In fine-tuning, we first train the 3D object detector on the large amount of labeled source domain data, and then further train and refine it with the small number of available labeled samples from the target domain. This setting of source and target domain data  has not been fully explored for domain adaptation in LiDAR-based 3D object detection. This data setting for domain adaptation is similar to ours. Therefore, we use fine-tuning as one of the baselines for comparison.

Lastly, our work is related to curriculum learning \cite{bengio2009curriculum} and a recent research in natural language processing \cite{xu2021gradual} that highlight the role of ordering in the training data for model learning. Their application is very different and they do not use the alternate training style. More concretely, we propose a training scheme for domain adaptation in LiDAR perception, specifically 3D object detection, where the training is performed using a batch of source and target samples in an alternate manner. Furthermore, the training data distribution gradually becomes similar to the target domain over the training since we keep reducing the amount of source data.

\section{Approach}
We aim to build a model that can adapt from a fully labeled source domain to an insufficiently labeled target domain for 3D object detection. In other words, we investigate a domain adaptation method for 3D object detection with partial supervision in the target domain. The objective is to leverage a fully labeled source domain and a partially labeled (and no unlabeled samples) target domain to train a 3D object detection model that makes the least possible error in its predictions on new samples (i.e., point clouds) from the target domain. 

Let us consider there is a source domain with a set of LiDAR point clouds and their labels that we denote as $D_s =\{x_i^s, y_i^s\}_{i=1}^{N_s}$. There is also a target domain with a set of limited number of labeled LiDAR point clouds $D_{tl} = \{x_i^t, y_i^t\}_{i=1}^{N_t}$ where $N_t << N_s$. Our goal is to train a domain adaptation model $M$ for 3D object detection on $D_s$ and $D_{tl}$, and then evaluate on unseen LiDAR point clouds from the target domain. We use the state-of-the-art 3D object detection model, CenterPoint \cite{yin2021center}, as our model for domain adaptation.

\begin{figure}[!ht] 
	\centering
	\includegraphics[width=0.45\textwidth, clip]{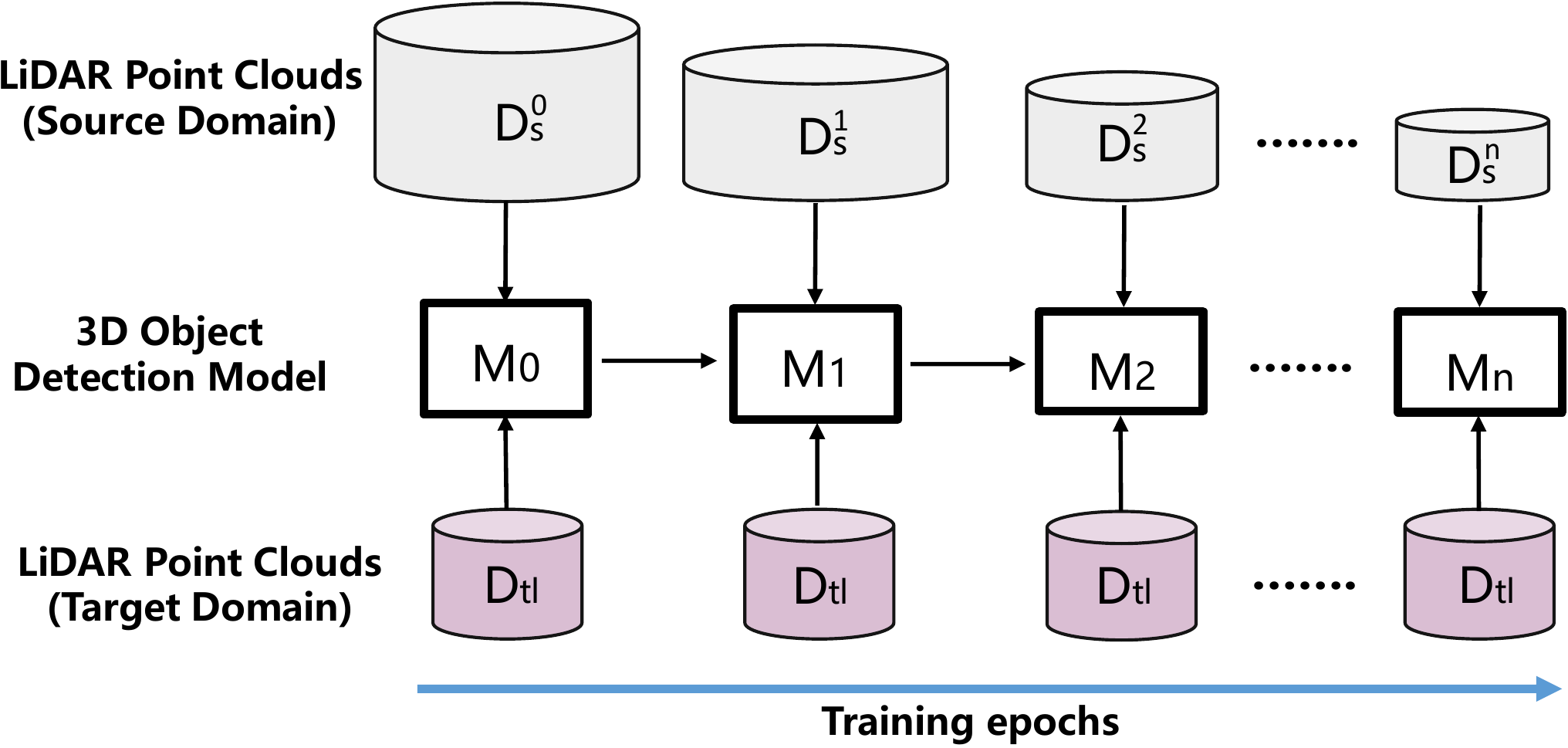} 
	\caption{Visualization of the Gradual Batch Alternation (\texttt{GBA}) training procedure. In each training step of an epoch, the model $M$ is trained by alternately using a batch of samples $\mathcal{B}_s$ from the source domain $D_s$ and a batch of samples $\mathcal{B}_t$ from the target domain $D_{tl}$. After every certain number of epochs $\mathcal{N}$ in the training, the amount of source domain data is reduced by a percent $\mathcal{P}$ than the preceding epochs (i.e., $D_s^i < D_s^{i-1}$), whereas the amount of target domain data is kept constant throughout.  The training data distribution slowly becomes similar to the target domain over time which allows the model $M$ to gradually fit to the target domain data distribution. }
	\label{fig:gba}
\end{figure}

A straightforward solution for domain adaptation in the setting discussed above is to make use of the fine-tuning strategy. In this strategy, CenterPoint can be pre-trained on the source domain data $D_s$. Next, this pre-trained model can be refined and trained until convergence on the small target domain data $D_{tl}$. The resulting CenterPoint model $M$ can then be used to evaluate on unseen LiDAR point clouds from the target domain for 3D object detection. We treat the fine-tuning strategy as one of the baselines in our experiments. In contrast to fine-tuning, where there the model abruptly deviates to the target domain from the source domain, we develop a progressive training method that allows the model to gradually shift to the target domain. We refer to this training technique as the Gradual Batch Alternation (\texttt{GBA}) training procedure which we describe in detail in the following section.

\subsection{Gradual Batch Alternation}\label{subsec:gba}
The core idea is to train the CenterPoint model $M$ using the source domain data $D_s$ and the target domain data $D_{tl}$ in such a manner that the model learns from both source and target initially but gradually pays more emphasis on learning from target.

We begin by training CenterPoint from scratch using the source and target data through supervised learning. We denote the initial model as $M_o$. At each training step, we alternately feed a batch of source samples ($\mathcal{B}_s \in D_s$) and a batch of target samples ($\mathcal{B}_t \in D_{tl}$) to the model. Note that we do not repeat the samples from a smaller domain to match the total number of samples from a larger domain in a training epoch. For instance, in the first training epoch, the source domain will have many more samples than the target domain. So, after some training steps of alternate training on the source and target samples, there will be no more samples left in the target domain $D_{tl}$. In this case, the training continues with each training step consisting of only the remaining samples from source $D_s$, and the first epoch ends when all the remaining samples from $D_s$ are used. 

Next, after every subsequent $\mathcal{N}$ epochs of training, we keep decreasing the amount of source domain data by a fixed percent $\mathcal{P}$ to obtain $D_s^i$ where $D_s^i < D_s^{i-1}$ , i.e., we gradually reduce the data from source domain, whereas keep the amount of target data constant over the training. We train the model $M_i$ on this reduced source $D_s^i$ and target $D_{tl}$ in the similar alternate fashion described above. As the training progresses, the training data distribution slowly moves closer to the target domain since  the data from source keeps decreasing. $\mathcal{N}$ and $\mathcal{P}$ are hyperparameters whose values are determined experimentally. We supervise the training using the losses described in CenterPoint \cite{yin2021center}. 

At the end of the $\texttt{GBA}$ training, we obtain the model $M$ which we use to evaluate on unseen samples (i.e., from validation set) from the target domain. 

Fig. \ref{fig:gba} illustrates the \texttt{GBA} training mechanism. We also summarize the \texttt{GBA} training in Algorithm \ref{algo:gba}. 

\begin{algorithm}[h] 
\begin{small}
\caption{Gradual Batch Alternation (GBA) Training}
\label{algo:gba}
\begin{algorithmic}[1]
\Require{source data $D_s^{0}$, target data $D_{tl}$, initial model $M_{0}$, epoch interval $\mathcal{N}$, source reduce percent $\mathcal{P}$, total epochs $\mathcal{E}$}
\Statex
\Function{GBA}{$D_s^{0}$, $D_{tl}$, $M_{0}$, $\mathcal{N}$, $\mathcal{P}$, $\mathcal{E}$}
    \State {$i$ $\gets$ $0$}
    \For{$n \gets 1$ to $\mathcal{E}$}
        \If{$n\mod \mathcal{N}$ is $0$}
            \State {$i, k \gets i+1, n$ floor division $\mathcal{N}$}
            \State {$D_s^{i}$ $\gets$ \Call{ReduceSource}{$D_s^{0}$, $k \times \mathcal{P}$}}
        \EndIf
        \State {$M_{n} \gets \Call{TrainAlternately}{M_{n-1}, D_s^{i}, D_{tl}}$}
    \EndFor
    \State \Return {$M_{n}$}
\EndFunction
\end{algorithmic}
\end{small}
\end{algorithm}

\section{Experiments}

\begin{table*}[t]
	\caption{Domain adaptation results for 3D object detection. We report results using AP\textsubscript{3D} of the Car category on the validation set of nuScenes \cite{caesar2020nuscenes}. We compare our method (\texttt{GBA}) with the several prior domain adaptation and baseline methods. We highlight the best domain adaptation results in bold.}
	\begin{threeparttable}
		\centering
		\renewcommand{\arraystretch}{1.3}
		\adjustbox{max width=\textwidth}{%
			\begin{tabular}{c|c|c|c}
				\hline
				\rotatebox{0}{Method} & 
				\rotatebox{0}{ONCE $\rightarrow$ nuScenes} & 
				\rotatebox{0}{PandaSet $\rightarrow$ nuScenes} &
				\rotatebox{0}{Waymo $\rightarrow$ nuScenes}  \\ \hline
				Centerpoint Oracle & 81.22 & 81.22 & 81.22 \\ \hdashline
				Mean Teachers \cite{tarvainen2017mean} & 49.00 & 53.50 & 53.00 \\
				SN \cite{wang2020train} & 38.20 & 48.70 & 43.70 \\
				Pseudo-labeling \cite{caine2021pseudo} & 47.80 & 55.00 & 51.30 \\
				ST3D \cite{yang2021st3d} & 49.48 & 55.10 & 52.67 \\ \hdashline
				  Source Only  & 47.45 & 48.77 & 47.52 \\
				Target Only  & 67.08 & 67.08 & 67.08 \\
				Fine-tuning  & 70.26 & 72.31 & 74.03 \\ \hdashline
				\texttt{GBA} (Ours) & \textbf{73.35} & \textbf{74.99} & \textbf{77.55} \\  
				\hline
			\end{tabular}%
		}
	\end{threeparttable}
	\label{table:main-results}
\end{table*}

\subsection{Settings}\label{sub:exp-settings}
{\bf Datasets.} We perform experiments with widely-adopted 3D LiDAR object detection datasets that have data captured from LiDAR sensors with different specifications. In particular, we experiment with ONCE \cite{mao2021once}, PandaSet \cite{xiao2021pandaset}, Waymo \cite{sun2020scalability}, and nuScenes \cite{caesar2020nuscenes} for domain adaptation.

ONCE \cite{mao2021once} is an autonomous driving dataset with $581$ sequences where scenes are captured using a LiDAR sensor that are transformed to dense 3D point clouds with around $70$K points in each sample. It has annotations for roughly $5$K samples in the training split, $3$K samples in validation split and $8$K samples in the testing split.

PandaSet \cite{xiao2021pandaset} contains $103$ sequences for 3D object detection with around $5$K samples in the training set and $3$K samples in the testing set.

Waymo \cite{sun2020scalability} Open Dataset is a large-scale autonomous driving dateset that provides $798$ labeled segments for training with around $158$K samples and $202$ labeled segments for validation with around $40$K samples.

nuScenes \cite{caesar2020nuscenes} is a popular LiDAR dataset with around $28$K and $6$K samples in training and validation, respectively.

\textbf{Source and Target Domains.} Our method aims to adapt from a label-rich source domain to an insufficiently labeled target domain. To simulate this scenario, we select ONCE, PandaSet and Waymo as the source domains and nuScenes as the target domain. From source domains, we only use their training splits. Furthermore, since we are interested in domain adaptation in the target domain with partial supervision, we randomly select $10$\% of samples and their labels from the training set of nuScenes for supervision. We refer to this data split as nuScenesDA10. Note that we discard the rest of the samples from nuScenes training set and do not use them in our method. We don't do unsupervised learning like prior methods, SN \cite{wang2020train}, Pseudo-labeling \cite{caine2021pseudo}, and ST3D \cite{yang2021st3d}.

{\bf Evaluation Metric.} We adopt the Average Precision metric (AP) defined in the official evaluation tool of the nuScenes dataset \cite{caesar2020nuscenes} for measuring the 3D object detection performance. Following prior work \cite{wang2020train}, we report the domain adaptation performance on the Car category.

{\bf Implementation Details.} We apply our domain adaptation technique to CenterPoint \cite{yin2021center} which is a state-of-the-art 3D object detection framework. We build our training method on top of OpenPCDet \cite{openpcdet2020}, a popular codebase for LiDAR-based 3D object detection. We follow the training settings in OpenPCDet for pre-training CenterPoint on a source domain wherever required. Following prior work \cite{yang2021st3d}, we fix the point cloud range to $[-75.2, 75.2]m$ for X and Y axes and $[-2, 4]m$ in Z-axis and set the voxel size to $[0.1, 0.1, 0.15]m$ for all the datasets. In our method \texttt{GBA}, we use the Adam \cite{kingma2014adam} optimizer with one cycle scheduler and set the initial learning rate to $5 \times 10^{-4}$. We fix the batch size to $48$ and train our models for $80$ epochs on four NVIDIA Tesla V100 32GB GPUs. In all the experiments, the hyperparameters $\mathcal{N}$ and $\mathcal{P}$ of $\texttt{GBA}$ are fixed to $18$ and $25$, respectively. We report the performance on the validation set of our target domain (i.e., nuScenes) using the last checkpoint from training.

{\bf Baseline and Comparison Methods.} We compare \texttt{GBA} against several baseline and prior methods. 

We define four baseline methods: $(i)$ Oracle, where a fully supervised model is trained on the target domain (i.e., the entire training set of nuScenes); $(ii)$ Source Only, where we directly evaluate a pre-trained source domain model on the target domain; $(iii)$ Target Only, where we train the model from scratch on the nuScenesDA10 split; $(iv)$ Fine-tuning, where we fine-tune a pre-trained model from a source domain on the nuScenesDA10 split. 

Since there is a lack of research in domain adaptation with partial supervision in the target domain for 3D object detection, we compare with state-of-the-art unsupervised domain adaptation (UDA) methods in images (Mean Teachers \cite{tarvainen2017mean}) and 3D object detection (Pseudo-labeling \cite{caine2021pseudo} and ST3D \cite{yang2021st3d}). We also compare with SN \cite{wang2020train}, a weakly supervised domain adaptation method for 3D object detection. Different from the UDA methods that utilize a large amount of unlabeled data from the target domain, we use a limited number of labeled samples from the target domain. Moreover, in order to perform a fair comparison, we re-implement these prior methods with CenterPoint \cite{yin2021center} as their backbone network. 

\subsection{Domain Adaptation Results}

In Table \ref{table:main-results}, we summarize and compare the performance our method \texttt{GBA} with the prior and baseline methods. \texttt{GBA} outperforms the prior domain adaptation methods for 3D object detection and several strong baselines by a large margin. By leveraging a small number of labeled samples from the target domain (i.e., nuScenesDA10), \texttt{GBA} achieves a very promising performance in comparison to the Oracle on nuScenes. More concretely, ONCE $\rightarrow$ nuScenes and PandaSet $\rightarrow$ nuScenes are just about $6$ to $8$\% inferior to Oracle. \texttt{GBA} further closes (less than $4$\%) the performance gap with Oracle on Waymo $\rightarrow$ nuScenes. These are very encouraging results indicating that \texttt{GBA} is a powerful domain adaptation method that can yield appealing 3D object detection results in a new target domain by capitalizing a limited number of labeled samples from that domain.

\begin{figure*}[!ht]
	\centering
	\begin{subfigure}[b]{0.3\textwidth}
		\centering
		\includegraphics[height=3.3cm,width=5.3cm]{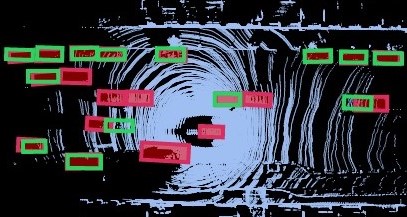}
		\caption{Mean Teachers \cite{tarvainen2017mean}}
		\label{fig:mean-teacher}
	\end{subfigure}
	\hfill
	\begin{subfigure}[b]{0.3\textwidth}
		\centering
		\includegraphics[height=3.3cm,width=5.3cm]{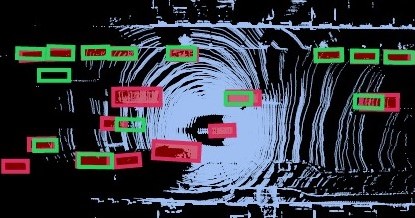}
		\caption{SN \cite{wang2020train}}
		\label{fig:sn}
	\end{subfigure}
	\hfill
	\begin{subfigure}[b]{0.3\textwidth}
		\centering
		\includegraphics[height=3.3cm,width=5.3cm]{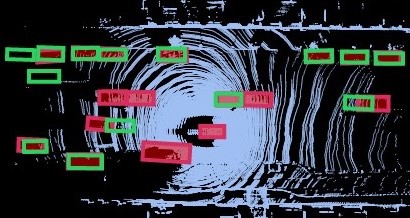}
		\caption{Pseudo-labeling \cite{caine2021pseudo}}
		\label{fig:pseudo-labeling}
	\end{subfigure}
	
	\begin{subfigure}[b]{0.3\textwidth}
		\centering
		\includegraphics[height=3.3cm,width=5.3cm]{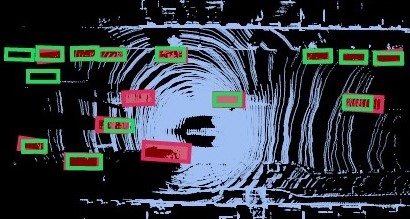}
		\caption{ST3D \cite{yang2021st3d}}
		\label{fig:st3d-waymo}
	\end{subfigure}
	\hfill
	\begin{subfigure}[b]{0.3\textwidth}
		\centering
		\includegraphics[height=3.3cm,width=5.3cm]{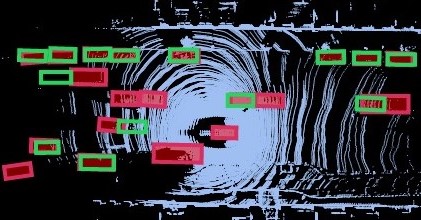}
		\caption{Source Only}
		\label{fig:source-only}
	\end{subfigure}
	\hfill
	\begin{subfigure}[b]{0.3\textwidth}
		\centering
		\includegraphics[height=3.3cm,width=5.3cm]{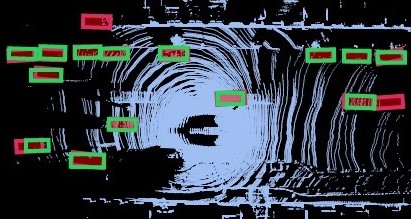}
		\caption{Target Only}
		\label{fig:target-only}
	\end{subfigure}
	
	\begin{subfigure}[b]{0.3\textwidth}
		\centering
		\includegraphics[height=3.3cm,width=5.3cm]{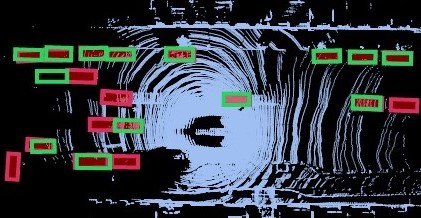}
		\caption{Fine-tuning}
		\label{fig:fine-tuning}
	\end{subfigure}
	\hfill
	\begin{subfigure}[b]{0.3\textwidth}
		\centering
		\includegraphics[height=3.3cm,width=5.3cm]{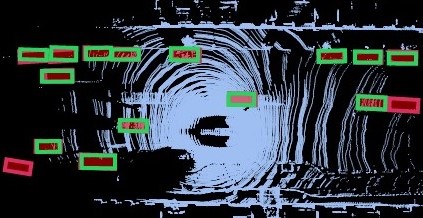}
		\caption{Oracle}
		\label{fig:oracle}
	\end{subfigure}
	\hfill
	\begin{subfigure}[b]{0.3\textwidth}
		\centering
		\includegraphics[height=3.3cm,width=5.3cm]{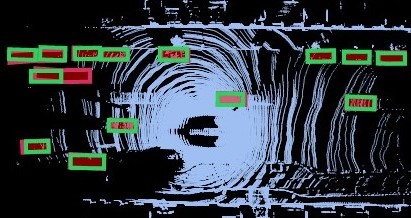}
		\caption{$\texttt{GBA}$ (Ours)}
		\label{fig:gf}
	\end{subfigure}
	\caption{Example qualitative results for Waymo $\rightarrow$ nuScenes. We visualize the predictions from the prior domain adaptation methods, baselines, Oracle, and our proposed method $\texttt{GBA}$ on a sample from nuScenes validation set in the bird's eye view (BEV). We show the point cloud in blue, predictions in red bounding boxes and ground-truth in green bounding boxes.}
	\label{fig:qual}
\end{figure*}

In Fig. \ref{fig:qual}, we show predictions from different methods on a point cloud from the nuScenes validation set for domain adaptation from Waymo $\rightarrow$ nuScenes. In contrast to the prior methods and baselines, $\texttt{GBA}$ (Fig. \ref{fig:gf}) produces less false positives (red bounding boxes) and achieves better overlap with the ground-truth (green bounding boxes). Moreover, it is very competitive to the fully supervised Oracle model (Fig. \ref{fig:oracle}). Thus, $\texttt{GBA}$ is a promising domain adaption technique.

\subsection{Ablation Study}
To understand the effectiveness of the \texttt{GBA} training, we conduct four additional experiments. 

First, we perform an experiment where we do not decrease the amount of source. In other words, we train the network with full source domain data and nuScenesDA10 by feeding their samples in alternate fashion in every epoch of the training. This is similar to prior work \cite{rist2019cross} except that we do not repeat the samples from the smaller target domain (i.e., nuScenesDA10) to match the training steps in an epoch with the large source domains. We denote this training scheme as vanilla batch alternation (\texttt{BA}). In Table \ref{table:ba-vs-gba}, we compare the domain adaptation results of \texttt{BA} and \texttt{GBA}. We find that \texttt{GBA} significantly improves over \texttt{BA} in domain adaptation.
\begin{table}[!h]
	\centering
	\caption{Comparison (in terms of AP\textsubscript{3D}) between \texttt{BA} and \texttt{GBA} on different domain adaptation tasks.}
	\begin{threeparttable}
		\renewcommand{\arraystretch}{1.3}
		\adjustbox{max width=0.48\textwidth}{%
			\begin{tabular}{c|c|c}
				\hline
				\rotatebox{0}{Method} & 
				\rotatebox{0}{$\texttt{BA}$} & 
				\rotatebox{0}{$\texttt{GBA}$}  \\ \hline
				ONCE $\rightarrow$ nuScenes & 64.55 & \textbf{73.35} \\
				PandaSet $\rightarrow$ nuScenes & 70.09 & \textbf{74.99} \\
				Waymo $\rightarrow$ nuScenes & 56.31 & \textbf{77.55} \\
				\hline
			\end{tabular}%
		}
	\end{threeparttable}
	\label{table:ba-vs-gba}
\end{table}

Second, we also study the impact of gradually decreasing the source domain data in $\texttt{GBA}$ training. This way we aim to identify whether reducing amount of source data after every $\mathcal{N}$ training epochs by percent $\mathcal{P}$ improves the 3D object detection performance on the target domain. Table \ref{table:source-amount} validates our hypothesis and shows improvement in the domain adaptation performance as the amount of source domain data is reduced during the training.
  
\begin{table}[!h]
	\centering
	\caption{Impact of reducing the amount of source domain data during the $\texttt{GBA}$ training. We show the performance (in AP\textsubscript{3D}) from different stages of the $\texttt{GBA}$ training. The training begins with full source domain data (i.e., $100\%$) but gradually the amount of source domain data is reduced as the training progresses.}
	\begin{threeparttable}
		\renewcommand{\arraystretch}{1.3}
		\adjustbox{max width=0.48\textwidth}{%
			\begin{tabular}{c|c|c|c|c|c}
				\hline
				\rotatebox{0}{Task} & 
				\rotatebox{0}{100\%} & 
				\rotatebox{0}{75\%} &
				\rotatebox{0}{50\%} &
				\rotatebox{0}{25\%} &
				\rotatebox{0}{0\%}  \\ \hline
				ONCE $\rightarrow$ nuScenes & 57.85 & 58.96 & 62.86 & 68.46 & 73.35 \\
				PandaSet $\rightarrow$ nuScenes & 60.98 & 		65.34 & 69.77 & 72.45 & 74.99 \\
				Waymo $\rightarrow$ nuScenes & 52.13 & 57.81 & 58.88 & 63.58 & 77.55 \\
				\hline
			\end{tabular}%
		}
	\end{threeparttable}
	\label{table:source-amount}
\end{table}

Third, we investigate how \texttt{GBA} performs on different domain adaptation tasks when less amount of labeled target data is used for partial supervision. Specifically, we prepare a smaller nuScenesDA5 split by randomly selecting $5$\% of samples and their labels from the training set of nuScenes. We run a strong baseline (fine-tuning) and our method \texttt{GBA} for domain adaptation using this split. Table \ref{table:da5-vs-da10} shows that the performance of both fine-tuning and \texttt{GBA} degrades with this split in comparison to the nuScenesDA10 split. However, \texttt{GBA} still outperforms fine-tuning and the prior domain adaptation methods (see their results in Table \ref{table:main-results}). Surprisingly, on Waymo $\rightarrow$ nuScenes, \texttt{GBA} with the nuScenesDA5 split is better than the fine-tuning with the nuScenesDA10 split. This implies that \texttt{GBA} can adapt to a new target domain and produce appealing results by utilizing even fewer labeled samples from the target domain.   
\begin{table}[!h]
	\centering
	\caption{Effect of using a smaller number of labeled samples from the target domain. Performance (in AP\textsubscript{3D}) of fine-tuning and \texttt{GBA} when adapting using a smaller nuScenesDA5 split. In the bracket, we also include the performance of fine-tuning and \texttt{GBA} using the nuScenesDA10 split from Table \ref{table:main-results} to assist in comparison.}
	\begin{threeparttable}
		\renewcommand{\arraystretch}{1.3}
		\adjustbox{max width=0.5\textwidth}{%
			\begin{tabular}{c|c|c}
				\hline
				\rotatebox{0}{Task} & 
				\rotatebox{0}{Fine-tuning} & 
				\rotatebox{0}{\texttt{GBA}} \\ \hline
				ONCE $\rightarrow$ nuScenes & 67.45 (70.26) & 68.77 (73.35) \\
				PandaSet $\rightarrow$ nuScenes & 69.73 (72.31)  & 69.66 (74.99) \\
				Waymo $\rightarrow$ nuScenes & 71.72 (74.03) & 74.22 (77.55) \\
				\hline
			\end{tabular}%
		}
	\end{threeparttable}
	\label{table:da5-vs-da10}
\end{table}

Lastly, we conduct an experiment where we replace CenterPoint with another state-of-the-art 3D detection model, PV-RCNN \cite{shi2020pv}. Table \ref{table:pv-rcnn-gba} compares the results between \texttt{GBA} and several strong baselines. Again, \texttt{GBA} achieves the best performance and outperforms the baselines by a margin. This implies that \texttt{GBA} is a model-agnostic method for domain adaptation in 3D object detection. 
\begin{table}[!h]
	\centering
	\caption{Domain adaptation results of different methods using the 3D object detection model, PV-RCNN \cite{shi2020pv}.}
	\begin{threeparttable}
		\renewcommand{\arraystretch}{1.3}
		\adjustbox{max width=0.5\textwidth}{%
			\begin{tabular}{c|c|c|c|c}
				\hline
				\rotatebox{0}{Task} & 
				\rotatebox{0}{Source Only} &
                \rotatebox{0}{Target Only} &
                \rotatebox{0}{Fine-tuning} &
				\rotatebox{0}{\texttt{GBA}} \\ \hline
				ONCE $\rightarrow$ nuScenes & 38.75 & 57.25 & 68.01 & \textbf{72.25}\\
				PandaSet $\rightarrow$ nuScenes & 46.90 & 57.25 & 69.93 & \textbf{72.02}\\
				Waymo $\rightarrow$ nuScenes & 43.97 & 57.25 & 69.24 & \textbf{73.18}\\
				\hline
			\end{tabular}%
		}
	\end{threeparttable}
	\label{table:pv-rcnn-gba}
\end{table}

\section{Conclusion}
In this paper, we investigated a less explored domain adaptation with partial supervision for LiDAR-based 3D object detection. We argued that this is a practical and optimistic setting for autonomous driving companies since our experiments showed that by introducing a limited number of labeled samples from the target domain, we are not only able to outperform the prior methods but also significantly reduce the gap with the fully supervised Oracle model in the target domain. We proposed the Gradual Batch Alternation (\texttt{GBA}) training pipeline that enables the model to steadily adapt to the target domain. Training of the model in $\texttt{GBA}$ begins with the plentiful source domain data and limited labeled target domain data, but with the progress in training less and less amount of source data is incorporated. We empirically showed that $\texttt{GBA}$ is a powerful domain adaptation technique for 3D object detection when compared to the prior methods and baselines. We hope this work will kindle interest in domain adaptation with partial supervision for perception systems in autonomous driving and that future research will adopt the proposed $\texttt{GBA}$ training process in other problems in this area when attempting to tackle the performance gap due to domain shifts.

\bibliographystyle{IEEEtran}
\bibliography{IEEEabrv,semida}


\end{document}